\documentclass{llncs} %

\usepackage{float}
\usepackage{graphicx}
\usepackage{wrapfig}

\usepackage[labelfont=bf]{caption}
\usepackage{color,soul}
\usepackage{amsmath} 
\usepackage{mathtools} 
\usepackage{amsfonts} 
\usepackage{amssymb}

\usepackage{array}
\usepackage{booktabs}
\usepackage{rotating}
\usepackage{multirow}
\usepackage{multicol}
\usepackage{lscape}
\usepackage{subfig}

\usepackage[export]{adjustbox}

\usepackage{algorithm,algorithmicx,algpseudocode}

\usepackage{color}

\usepackage{color}

\definecolor{babyblue}{rgb}{0.54, 0.81, 0.94}
\definecolor{armygreen}{rgb}{0.29, 0.33, 0.13}
\definecolor{brightlavender}{rgb}{0.75, 0.58, 0.89}
\definecolor{aqua}{rgb}{0.0, 1.0, 1.0}
\definecolor{caribbeangreen}{rgb}{0.0, 0.8, 0.6}
\definecolor{reddish}{rgb}{0.82, 0.1, 0.26}
\definecolor{caribbeangreen}{rgb}{0.31, 0.78, 0.47}
\definecolor{jasper}{rgb}{0.84, 0.23, 0.24}
\definecolor{red}{rgb}{1.0, 0.0, 0.0}
\definecolor{green}{rgb}{0.0, 1.0, 0.0}
\definecolor{blue}{rgb}{0.0, 0.0, 1.0}
\definecolor{darkgreen}{rgb}{0.1, 0.7, 0.1}
\definecolor{darkblue}{rgb}{0.1, 0.1, 0.7}
\definecolor{red}{rgb}{0.7, 0.1, 0.1}
\definecolor{coral}{rgb}{1.0, 0.5, 0.31}

\usepackage[colorlinks=true,linkcolor=coral,citecolor=coral]{hyperref}
\usepackage[colorinlistoftodos,prependcaption,textsize=tiny]{todonotes}

\usepackage{tikz,xcolor,hyperref}

\definecolor{lime}{HTML}{A6CE39}
\DeclareRobustCommand{\orcidicon}{
	\begin{tikzpicture}
	\draw[lime, fill=lime] (0,0) 
	circle [radius=0.16] 
	node[white] {{\fontfamily{qag}\selectfont \tiny ID}};
	\draw[white, fill=white] (-0.0625,0.095) 
	circle [radius=0.007];
	\end{tikzpicture}
	\hspace{-2mm}
}

\foreach \x in {A, ..., Z}{\expandafter\xdef\csname orcid\x\endcsname{\noexpand\href{https://orcid.org/\csname orcidauthor\x\endcsname}
			{\noexpand\orcidicon}}
}
			
\begin{document}

\title{Multi-View Brain HyperConnectome AutoEncoder For Brain State Classification}

\titlerunning{Short Title}  

\author{Alin Banka\index{Banka, Alin}  \and Inis Buzi\index{Buzi, Inis}  \and Islem Rekik\orcidA{} \index{Rekik, Islem}\thanks{ {corresponding author: irekik@itu.edu.tr, \url{http://basira-lab.com}, GitHub: \url{http://github.com/basiralab/HCAE}. }}}

\institute{BASIRA Lab, Faculty of Computer and Informatics, Istanbul Technical University, Istanbul, Turkey}

\authorrunning{A Banka et al.}

\maketitle              

\begin{abstract}

Graph embedding is a powerful method to represent graph neurological data (e.g., brain connectomes) in a low dimensional space for brain connectivity mapping, prediction and classification. However, existing embedding algorithms have two major limitations. \emph{First}, they primarily focus on preserving \emph{one-to-one} topological relationships between nodes (i.e., regions of interest (ROIs) in a connectome), but they have mostly ignored \emph{many-to-many} relationships (i.e., set to set), which can be captured using a \emph{hyperconnectome} structure. \emph{Second}, existing graph embedding techniques cannot be easily adapted to \emph{multi-view} graph data with heterogeneous distributions. In this paper, while cross-pollinating adversarial deep learning with hypergraph theory, we aim to \emph{jointly} learn deep latent embeddings of subject-specific  multi-view brain graphs to eventually disentangle different brain states such as Alzheimer's disease (AD) versus mild cognitive impairment (MCI). First, we propose a new simple strategy to build a hyperconnectome for each brain view based on nearest neighbour algorithm to preserve the connectivities across pairs of ROIs. Second, we design a hyperconnectome autoencoder (HCAE) framework which operates directly on the multi-view hyperconnectomes based on hypergraph convolutional layers to better capture the many-to-many relationships between brain regions (i.e., graph nodes). For each subject, we further regularize the hypergraph autoencoding by adversarial regularization to align the distribution of the learned hyperconnectome embeddings with the original hyperconnectome distribution. We formalize our hyperconnectome embedding within a geometric deep learning framework to optimize for a given subject, thereby designing an \emph{individual-based} learning framework.  Our experiments showed that the learned embeddings by HCAE yield to better results for AD/MCI classification compared with deep graph-based autoencoding methods. Our HCAE code is available in Python at \url{http://github.com/basiralab/HCAE}.

\end{abstract}


\section{Introduction}

Magnetic resonance imaging (MRI) has introduced exciting new opportunities for understanding the brain as a complex system of interacting units in both health and disease and across the human lifespan. Based on MRI data, the brain can be represented as a connectomic graph (i.e., connectome), where the connection between different anatomical regions of interest (ROIs) is modeled. A large body of research work showed how the brain connectome gets altered by neurological disorders, such as mild cognitive impairment (MCI) or Alzheimer's disease (AD) \cite{Fornito:2015,Heuvel:2019}. In network neuroscience, the brain connectome, which is typically regarded as a graph encoding the low-order \emph{one-to-one} relationships between pairs of ROIs, presents a macroscale representation of the interactions between anatomical regions at functional, structural or morphological levels \cite{Bassett:2017}. Analyses of either network connectivity or topology for examining variations in the type and strength of connectivity between brain regions, are integral to understanding the connectomics of brain disorders \cite{Fornito:2015} as well as the cross-disorder connectome landscape of brain dysconnectivity \cite{Heuvel:2019}. Although graphs enable a powerful interactional representation of the data, they present a reductionist representation of the brain complexity with their simple topology where edges connect at max two nodes. Hence, this limits the learning of representations for complex interactions between brain regions in tasks such as connectome classification, connectomic disease propagation prediction, and holistic brain mapping. 

To address this first gap in the connectomics literature \cite{Fornito:2015,Bassett:2017,Heuvel:2019,Bullmore:2009} while drawing inspiration from the hypergraph theory, we present the concept of the brain \emph{hyperconnectome}, which models the high-order \emph{many-to-many} interactions between brain regions. Specifically, in the hyperconnectome, one naturally captures the relationships between \emph{sets} of ROIs, where a hyperedge can link more than two nodes. In other words, the hyperconnectome permits us to overcome the limitations of one-to-one (i.e., low-order) relationships between ROIs in traditional graph-based connectomes \cite{Bullmore:2009,Heuvel:2019} by introducing many-to-many (i.e., high-order) relationships between nodes. Although compelling, such high-order brain representation introduces more challenges to machine learning in connectomics as it increases the dimensionality of the data. To address the curse of dimensionality, this brings us to deploying dimensionality reduction or data embeddings, which are pivotal for learning-based tasks such as classification between healthy and disordered brains or unsupervised clustering of brain states. However, traditional machine learning techniques such as principle component analysis (PCA) fail to preserve the \emph{topological relationships} between the nodes of the brain connectomes as they operate on vectorized brain connectomes. Recently, the emerging field of geometric deep learning \cite{Bronstein:2017,Zhang:2018,Wu:2019,Zhou:2018}, aiming to adapt deep learning on Euclidean data such as images to non-Euclidean data including graphs and manifolds, has allowed us not only to generalize deep learning on graphs but also to learn latent representations of graphs in a low-dimensional space. More recently, \cite{banka} proposed adversarial connectome embedding (ACE) based on graph convolutional networks \cite{gcn}, which preserves (i) the topological structures of the brain connectomes when generating low-dimensional embeddings, and (ii) enforces the distribution of the learned connectome embeddings to match that of the orginial connectomes. Although promising, such approaches are not naturally designed to operate on hypergraphs, let alone \emph{multi-view} hypergraphs where each hypergraph encodes the high-order relationship between nodes using a particular data view (e.g., function or morphology in hyperconnectomes). Exceptionally, a recent work \cite{hgnn} cross-pollinated hypergraph theory and deep learning to introduce hypergraph neural networks (HGNN). However, this was primarily designed within a transductive learning setting where inference on hypergraph aims to minimize the label difference among nodes with stronger connections on hypergraph to eventually assign labels to unlabelled nodes. As such, one can learn how to label nodes of testing samples in a hypergraph. However, this stands out from fully unsupervised data autoencoding, where the primary focus is on learning meaningful and representative embeddings of the input data by minimizing its self-reconstruction loss. To the best of our knowledge, hypergraph autoencoding networks are currently absent in the state-of-the-art.

To address these limitations, we propose the first geometric hypergraph autoencoder, named HyperConnectome AutoEncoder (HCAE), for embedding  multi-view brain hyperconnectomes derived from multi-view brain connectomes where each individual is represented by a set of brain connectomes, each capturing a particular \emph{view of the brain} (e.g., sulcal depth dissimilarity). \emph{First}, we introduce the definition of the hyperconnectome structure. \emph{Second}, we leverage hypergraph convolutional layers introduced in HGNN \cite{hgnn} to design our HCAE architecture, which preserves the hyperconnectome relationships when learning the latent embeddings. Unlike existing typical graph autoencoders \cite{banka,arga} which are trained on a population of samples, we formalize our problem as a \emph{subject-specific} loss function to optimize for each individual hyperconnectome. For each subject, to generate a hyperconnectome embedding which is true to the input multi-view brain networks, we further introduce a hyperconnectome adversarial regularization, where we integrate a discriminator network to force the latent representations to match the prior distribution of the subject-specific multi-view hyperconnectome.  \emph{Third}, we demonstrate the utility of our architecture in integrating multi-view hyperconnectome by learning a shared embedding exploiting the cross-view relationship between diverse hyperconnectome representation of the same brain. Ultimately, we evaluate the discriminative power of the learned embeddings in distinguishing between different brain states by training a support vector machine classifier using repeated randomized data partition for reproducibility.

\begin{figure}[htp!]
\centering
\hspace*{-0.7cm}
\includegraphics[width=14cm]{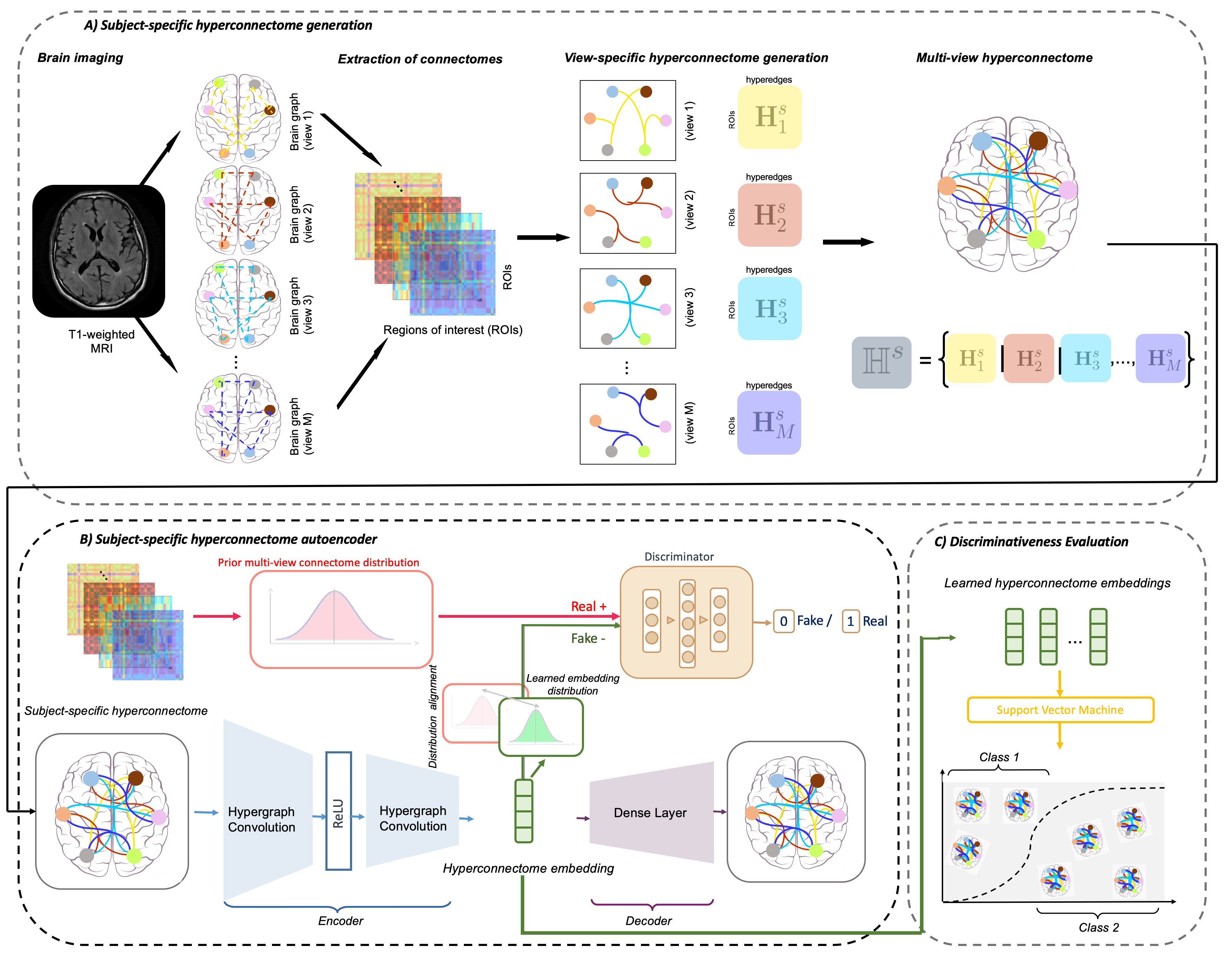}
\caption{\emph{Hyperconnectome Autoencoder (HCAE) architecture for multi-view brain network state classification.}  \textbf{(A)} For each subject, we generate a set of multi-view brain connectomes, each capturing a connectional view of the brain (e.g., function or morphology). From each brain connectome, we construct a view-specific hyperconnectome which consists of hyperedges, built by connecting each node (i.e., an anatomical region of interest (ROI)) to its $k$-nearest neighboring nodes. Ultimately, we stack horizontally the incidence matrices from separate views to create a \emph{multi-view brain hyperconnectome}, which captures the \emph{high-order} connectivity from different complementary brain views. \textbf{(B)} HCAE is trained in a \emph{subject-specific} manner, i.e., for each individual independently. The architecture consists of three main components: (1) the encoder, (2) the decoder, and (3) the discriminator. The encoder, consisting of 2 hypergraph convolutional layers and a ReLU activation function in between the layers, embeds the input multi-view hyperconnectome to a low-dimensional space. The decoder reconstructs the subject-specific multi-view brain hyperconnectome from the learned embedding through a single dense layer.  The discriminator acts as a regularizer where it enforces the distribution of the learned hyperconnectome embeddings to match that of the original hyperconnectome distribution in an adversarial manner. \textbf{(C)} Evaluating the discriminativeness of the learned embeddings in distinguishing between brain states using support vector machines (SVM). We use the encoded multi-view hyperconnectomes to train an SVM classifier to classify two brain states (e.g., AD and MCI).} 
\label{fig:1}
\end{figure}

\section{Proposed HyperConnectome AutoEncoder (HCAE) for Brain State Classification}

\textbf{Problem statement}. Let $\mathcal{G}^s = \{\mathbf{G}_1,\dots,\mathbf{G}_M \}$ denote a set of multi-view brain connectomes of subject $s$ in the population $\mathbb{G}$, comprising $M$ fully-connected brain graphs where $\mathbf{G}_m$ represents the brain graph derived from measurement $m$ (e.g., correlation in neural activity or similarity in morphology). Each brain graph $\mathbf{G}_m = (V,E, \mathbf{X}_m) \in \mathcal{G}^s$ captures a \emph{connectional view} of the brain wiring, where $V$ denotes a set of $N$ brain ROIs, $E$ a set of edges connecting pairs of nodes, and $\mathbf{X}_m \in \mathbb{R}^{N \times N}$ denotes a symmetric brain connectivity matrix encoding the pairwise relationship between brain ROIs. Our goal is to learn \emph{discriminative} and \emph{representative} latent representation of each individual $\mathcal{G}^s$ by capitalizing on hypergraph structure and the nascent field of geometric deep learning.

\textbf{A-	Multi-view hyperconnectome definition.} Given a brain graph $\mathbf{G}_m = (V,E, \mathbf{X}_m) \in \mathcal{G}^s$ of subject $s$, we define a hyperconnectome of view $m$ $\mathcal{H}^s_m = \{ V, \mathbf{\mathcal{E}_m}, \mathbf{W}_m^s\}$ as a set of nodes $V$ and a set of hyperedges $\mathbf{\mathcal{E}_m}$, where each hyperedge is assigned a unitary weight in the diagonal weight matrix $\mathbf{W}_m \in \mathbb{R}^{|\mathcal{E}| \times |\mathcal{E}|}$. Basically, we construct the hypergraph $\mathcal{H}^s_m$ by connecting each node $v$ in $V$ to its $k$-nearest neighboring nodes, thereby defining a hyperedge $e \in \mathbf{\mathcal{E}_m}$. Next, we create a view-specific hyperconnectome incidence matrix $\mathbf{H}_m^s \in  \mathbb{R}^{|V|  \times |\mathcal{E}|}$, the entries of which are defined as:

\begin{equation}
  \mathbf{H}_m^s(e, v)=\begin{cases}
    1, & \text{if $v \in e$}.\\
    0, & \text{if $v \notin e$}.
  \end{cases}
\end{equation}
for $v \in V$ and $e \in \mathcal{E}$.  

Additionally, each node $v_i$ in $\mathcal{H}^s_m$ stores a set of features, which represent the $i^{th}$ row in the graph connectivity matrix $\mathbf{X}_m^s$ to preserve the relationship between different nodes since our incidence matrix is binary. Ultimately, by constructing a hyperconnectome for each view $m$, we create a multi-view hyperconnectome $\mathbb{H}^s$ stacking horizontally incidence matrices from different views (\textbf{Fig.}~\ref{fig:1}):

\begin{equation*}
    \mathbb{H}^s = \{
    \mathbf{H}_{1}^s|\mathbf{H}_{2}^s|, \dots, |\mathbf{H}_{M}^s
    \},
\end{equation*}

In parallel, we also stack the connectivity matrices of all views as follows:

\begin{equation*}
    \mathbb{X}^s = \{
    \mathbf{X}_{1}^s|\mathbf{X}_{2}^s|, \dots,|\mathbf{X}_{M}^s
    \}
\end{equation*}

\textbf{B-	Subject-specific hyperconnectome autoencoder.}
Next, we construct and train the HCAE model in order to extract low dimensional representations of our hyperconnectome. The model consists of three main components (\textbf{Fig.}~\ref{fig:1}): (1) the encoder constructed using hypergraph convolutional layers, (2) the decoder constructed using dense layers, and (3) the discriminator, which is utilized for adversarial distribution regularization, and composed of dense layers. The encoder consists of two stacked hypergraph convolutional layers. The two layers utilize the traditional hypergraph convolutional operation \cite{hgnn}. A rectified linear unit (ReLu) activation function is used after the first layer, and a linear activation function is added after the second one.

\begin{gather*}
    \mathbf{Y}^{(l)} = \phi(\mathbf{\mathbf{D}_{v}}^{-1/2}\mathbb{H}^s  \mathbf{W}_m \mathbf{\mathbf{D}_{e}}^{-1}(\mathbb{H}^s)^{T}   \mathbf{\mathbf{D}_{v}}^{-1/2}\mathbf{Y}^{(l-1)}\mathbf{\Theta}^{(l)}),
\end{gather*}

where $d(v)=\sum_{\mathbf{e} \in \mathcal{E}} \mathbf{W}_{m}\mathbb{H}^s{(v,e)}$ and $\delta(v)=\sum_{\mathbf{v} \in V}\mathbb{H}^s{(v,e)}$ are respectively definitions of the vertex degree and edge degree. $\mathbf{D}_{v} \in \mathbb{R}^{V \times V}$ and $\mathbf{D}_{e} \in \mathbb{R}^{\mathcal{|E|} \times \mathcal{|E|}}$ are respectively the diagonal matrices of the vertex degrees and edge degrees. $\mathbf{W}_m$ is the hyperedge weight matrix, $\mathbf{\Theta}^{(l)}$ represents the learned filter (i.e., learned weights of the hypergraph convolutional layers) applied to all hyerconnectome nodes to extract high-order connectional features, and $\mathbf{Y}^{(l-1)}$ represents the input produced by the previous layer. $\mathbf{Y}^{(0)}$ = $\mathbb{X}^s$ denotes the \emph{multi-view} brain connectivity matrix of the subject, and $\phi$ represents the activation function. The low-dimensional latent embedding is obtained as the output of the second convolutional layer, $\mathbf{Z}$ = $\mathbf{Y}^{(2)} \in \mathbb{R}^{N}$. Notice that HCAE convolutions are hypergraph convolutions. The purpose of the hypergraph convolutional layer is to take advantage of the high-order correlation among data \cite{hgnn}. It does so by incorporating the node-edge-node transformation. Specifically, the $d$-dimensional feature is extracted by processing the node feature matrix  $\mathbf{Y}^{(0)}$ with the learnable weight matrix $\mathbf{\Theta}^{(l)}$. The output extracted is gathered to construct the hyperedge feature through multiplication with $(\mathbb{H}^s)^{T}$. Finally, the output is received through hyperedge feature aggregation, achieved by multiplying with $\mathbb{H}^s$. $\mathbf{D}_{v}$ and $\mathbf{D}_{e}$ are included in the equation for normalization. The reconstruction of the hyperconnectome is done through a dense layer, $\mathbb{\widetilde{H}}^{s}$ = $\mathbf{Z}  \mathcal{W}$, where $\mathcal{W}$ denotes the learned weights of the dense reconstruction layer, and $\mathbb{\widetilde{H}}$ is the reconstructed incidence matrix. The loss is defined as follows:

\begin{gather}
    L_{0} = E_{q(\mathbf{Z}|\mathbb{X}^s,\mathbb{H}^s)}[\log P(\mathbb{\widetilde{H}}^{s} | \mathbf{Z})]
\end{gather} 

Drawing inspiration from \cite{arga}, we utilize adversarial regularization to force the low-dimensional latent embeddings to match the distribution of the input multi-view brain networks (\textbf{Fig.}~\ref{fig:1}). Specifically, we integrate an adversarial discriminator as a multilayer perceptron (MLP) comprising multiple stacked dense layers. A discriminator $D$, which is primarily employed in Generative Adversarial Networks (GANs), is commonly trained to discriminate between real samples and fake samples generated by another network known as the generator $G$. In our case, we integrate the loss of the discriminator into the model in order to force the learned low-dimensional embedding of a hyperconnectome to align better with the prior distribution of the input multi-view brain networks of a single subject:

\begin{gather}
    -\frac{1}{2}E_{\mathbf{z} \sim p_{Z}}[\log{D}(\mathbf{Z})] - 
    \frac{1}{2}E_{\mathbb{X}^s}[\log{(1 - D(G(\mathbb{X}^s,\mathbb{H}^s)))}]
\end{gather}

As such, we create a form of regularization that propels a better autoencoding by solving the following min-max optimization problem:

\begin{gather}
    \min_{G} \max_{D} E_{\mathbf{Z} \sim p_{Z}}[\log{D}(\mathbf{Z})] + 
    E_{\mathbf{Z} \sim p_{Z}}[\log{(1 - D(G(\mathbb{X}^s,\mathbb{H}^s)))}]
\end{gather}

\section{Results and Discussion}

\textbf{Dataset, parameters, and benchmarking.}
In our experiments, we evaluated our model with the ADNI GO public dataset \cite{mueller} over 77 subjects (36 MCI and 41 AD), each with structural T1-w MR image. Each subject has 4 cortical morphological networks derived as explained in \cite{Mahjoub:2018,Dhifallah:2020,Nebli:2019}. Each cortical morphological network is encoded in  a symmetric adjacency matrix with size ($35 \times 35$), generated using a specific cortical attribute: (1) maximum principal curvature, (2) cortical thickness,  (3) sulcal depth, and (4) average curvature. We compared our proposed framework with Adversarial Connectome Embedding (ACE) \cite{banka}, a geometric autoencoder architecture, which captures the region relationships \emph{one-to-one} in a graph structure. It utilizes conventional graph convolutional layers to learn the embeddings and adversarial regularizing network for original-encoded distribution alignment. We evaluated both ACE and HCAE on single-view and multi-view connectomes. Since ACE \cite{banka} does not naturally operate on multi-view data, we averaged the multi-view connectomes for evaluating it on multi-view data. HCAE is trained on each subject, independently, using 30 epochs. 

\begin{figure}[htp!]
\centering
{\includegraphics[width=11.5cm]{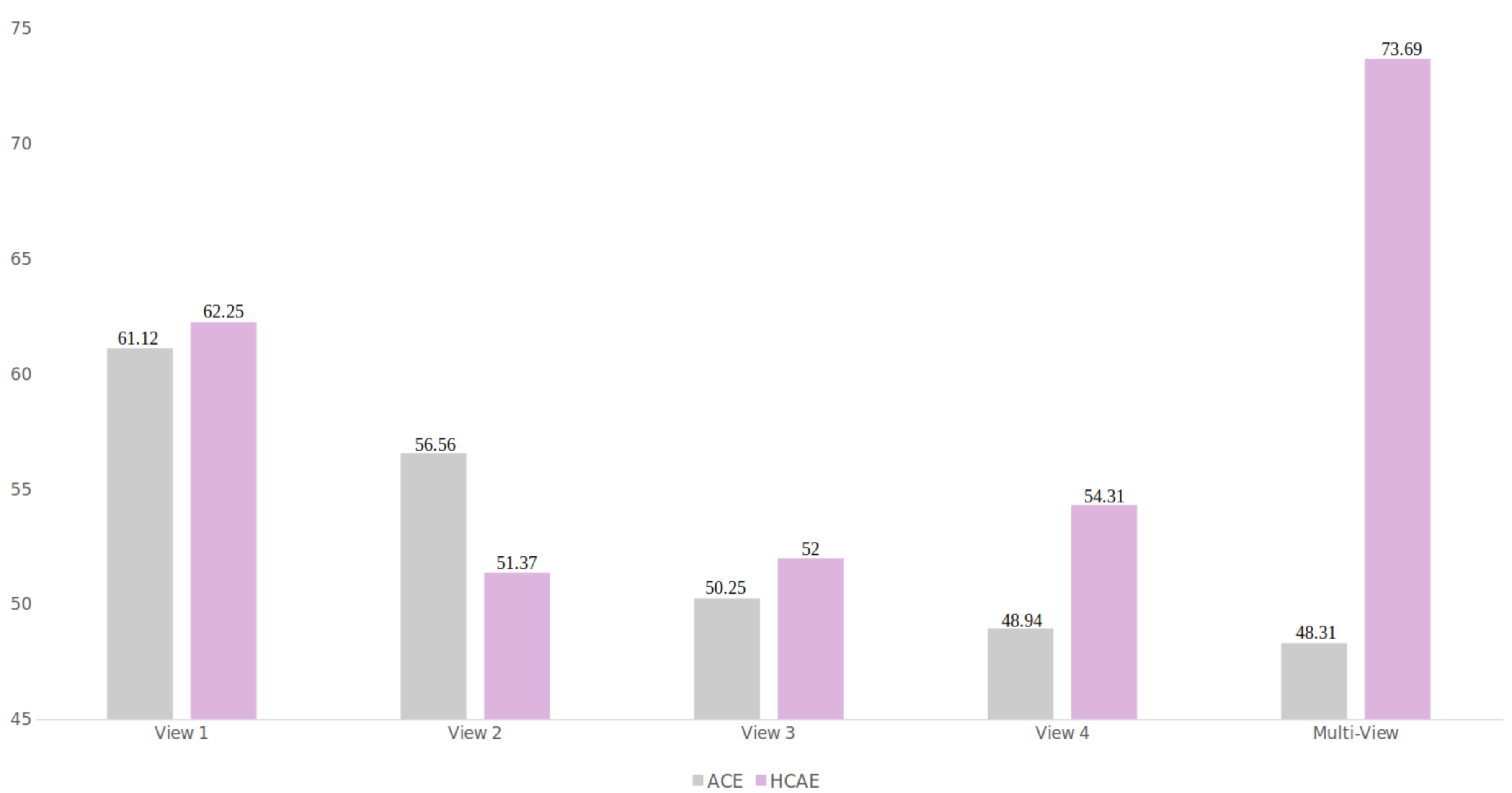}}
\caption{\emph{Comparison of SVM classification accuracies of AD and MCI brain states using low-order embeddings learned by ACE \cite{banka} and high-order embeddings learned by our HCAE.}}
\label{fig:2}
\end{figure}
 
\begin{figure}[h!]
\centering
{\includegraphics[width=11.5cm]{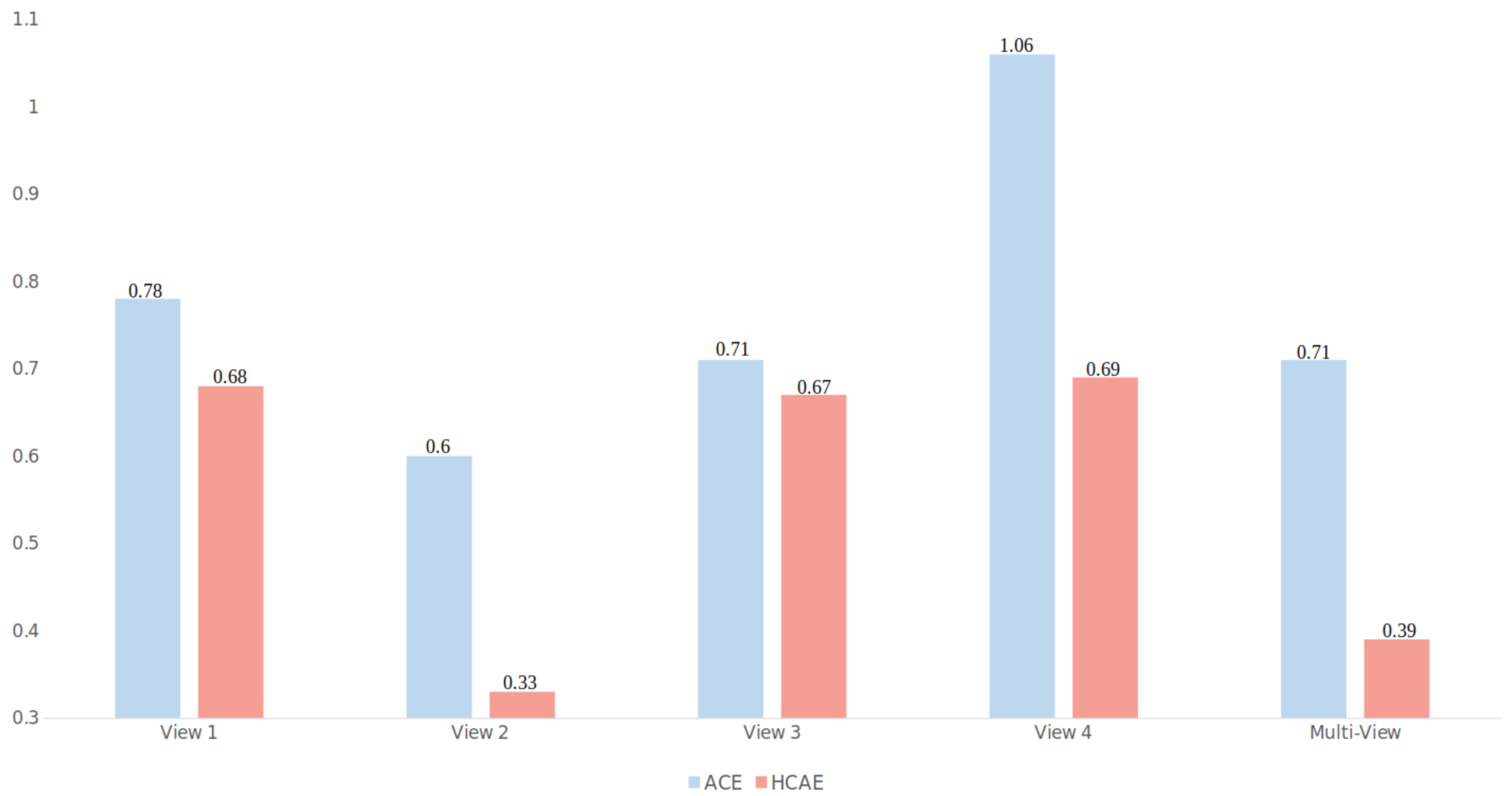}}
\caption{\emph{Comparison of hyperconnectome reconstruction errors by ACE \cite{banka} and HCAE.}}
\label{fig:3}
\end{figure}

\textbf{Evaluation of the discriminativeness of the learned hyperconnectome embeddings.}
We randomly split our dataset into 80\% training samples and  20\% testing subjects to train and test an SVM classifier using the learned high-order embeddings by ACE and HCAE. Average classification results over 100 repeated runs are reported in \textbf{Fig.}~\ref{fig:2} and cross-entropy reconstruction error in \textbf{Fig.}~\ref{fig:3}. Remarkably, the AD/MCI classification accuracy was largely boosted from 48.31\% to 73.69\% when using the multi-view representation of the brain network in comparison with  single-view connectomes. Excluding view 2, HCAE achieved the best results. Furthermore, the data reconstruction error was much lower using our HCAE model on both single and multi-view connectomic datasets. This demonstrates that our model better preserves the high-order brain connectivity topology.

\textbf{Limitations and recommendations for future work.}
Even though our model produced the best results in brain state classification and hyperconnectome autoencoding in both single and multiple brain views, it has a few limitations that could be overcome in future work.  \emph{First}, the learned embeddings are learned in a fully unsupervised manner, one possible extension is to integrate a second discriminator to learn \emph{discriminative} embeddings supervised by the input brain states. As such, one needs to adapt HCAE to population-driven training.  \emph{Second}, in our model we leverage multi-channel multiplication to operate on a  multi-view hyperconnectome. Alternatively, one can add a hyperconnectome fusion block to integrate  hyperconnectomes in weighted manner across all views.  \emph{Finally}, although our HCAE nicely learns the embeddings capturing many-to-many relationships between brain regions, which was demonstrated to be useful for brain state classification, it cannot handle spatiotemporal multi-view brain connectomes, which are time-dependent. As a future extension of our HCAE, we aim to extend to autoencode evolution trajectories of multi-view brain connectomes leveraging recurrent neural networks in geometric deep learning \cite{jain2016structural,hajiramezanali2019variational} to model the temporal relationships between connectomes acquired at different timepoints.

\section{Conclusion}
In this paper, we proposed the first hyperconnectome autoencoder framework for brain state classification. Our HCAE operates on multiple brain connectomes, leverages adversarial learning for hyperconnectome autoencoding using hypergraph convolutional layers, and learns representative high-order representation of brain connectivity that was shown to be discriminative in our experiments. This work develops the field of network neuroscience along the hyperconnectivity front, which aims to present a holistic representation of different facets of the brain connectome. This further calls for more scalable applications of the proposed multi-view hyperconnectome representation to unify our understanding of brain structure, function, and morphology, and how these get altered in a wide spectrum of brain disorders.

\section{Supplementary material}

We provide two supplementary items for reproducible and open science:

\begin{enumerate}
	\item A 6-mn YouTube video explaining how our prediction framework works on BASIRA YouTube channel at \url{https://youtu.be/ncPyj_4cSe8}.
	\item An improved version of the adversarial brain multiplex generation code is available on GitHub at \url{https://github.com/basiralab/HCAE}. 
\end{enumerate}

\section{Acknowledgments}

I. Rekik is supported by the European Union's Horizon 2020 research and innovation programme under the Marie Sklodowska-Curie Individual Fellowship grant agreement No 101003403 (\url{http://basira-lab.com/normnets/}).


\bibliography{Biblio3}
\bibliographystyle{splncs}
\end{document}